\documentclass{article}
\usepackage[margin=1.0in]{geometry}

\usepackage{graphicx} 
\usepackage{amsmath}
\usepackage{amssymb}
\usepackage{subcaption}
\usepackage{graphicx}
\usepackage{url}
\usepackage{color, colortbl}
\usepackage{csquotes}
\usepackage{booktabs}
\usepackage{float}
\usepackage{bm}
\usepackage{bbm}
\usepackage{multirow}
\usepackage{tabularx}
\usepackage[inline]{enumitem}
\usepackage{soul}
\newcommand{\ie}{\textit{i.e.}, }
\newcommand{\eg}{\textit{e.g.}, }
\newcommand{\etal}{\textit{et al.}}

\DeclareCaptionFormat{custom}
{%
    \textbf{#1#2}\textit{\small #3}
}
\makeatletter
\DeclareRobustCommand{\change}{%
  \@bsphack
  \normalcolor 
  \@esphack
}
\DeclareRobustCommand{\stopchange}{%
  \@bsphack
  \normalcolor
  \@esphack
}
\makeatother

\title{ \bf
Diver Interest via Pointing in Three Dimensions:

3D Pointing \change Reconstruction \stopchange for Diver-AUV Communication*
}
\date{}

\author{Chelsey Edge$^{1}$, Demetrious Kutzke$^{1}$, Megdalia Bromhal$^{2}$, and Junaed Sattar$^{1}$%
\thanks{*This work was supported in part by the National Science Foundation Grant IIS-2220956 and the Science, Mathematics, and Research for Transformation (SMART) Program.}
\thanks{$^{1}$Chelsey Edge, Demetrious Kutzke, and Junaed Sattar are with the Department of Computer Science \& Engineering, University of Minnesota--Twin Cities, Minneapolis, MN 55355, USA
        {\tt\small \{edge0037,kutzk015,junaed\}@umn.edu}}%
\thanks{$^{2}$Megdalia Bromhal is with the Department of Computer Science, University of North Carolina Wilmington
        {\tt\small mbromhal@umn.edu}}%
}

\begin{document}

\maketitle

\begin{abstract}

This paper presents Diver Interest via Pointing in Three Dimensions (DIP-3D), a method to relay an object of interest from a diver to an autonomous underwater vehicle (AUV) by pointing that includes three-dimensional distance information to discriminate between multiple objects in the AUV's camera image. Traditional dense stereo vision for distance estimation underwater is challenging because of the relative lack of saliency of scene features and degraded lighting conditions. Yet, including distance information is necessary for robotic perception of diver pointing when multiple objects appear within the robot's image plane. We subvert the challenges of underwater distance estimation by using sparse reconstruction of keypoints to perform pose estimation on both the left and right images from the robot's stereo camera. Triangulated pose keypoints, along with a classical object detection method, enable DIP-3D to infer the location of an object of interest when multiple objects are in the AUV's field of view. By allowing the scuba diver to point at an arbitrary object of interest and enabling the AUV to autonomously decide which object the diver is pointing to, this method will permit more natural interaction between AUVs and human scuba divers in underwater-human robot collaborative tasks. 

\end{abstract}
\section{Introduction}
\label{sec:introduction}


\begin{figure}
\vspace{1mm}

    \centering
        \includegraphics[width=.98\linewidth]{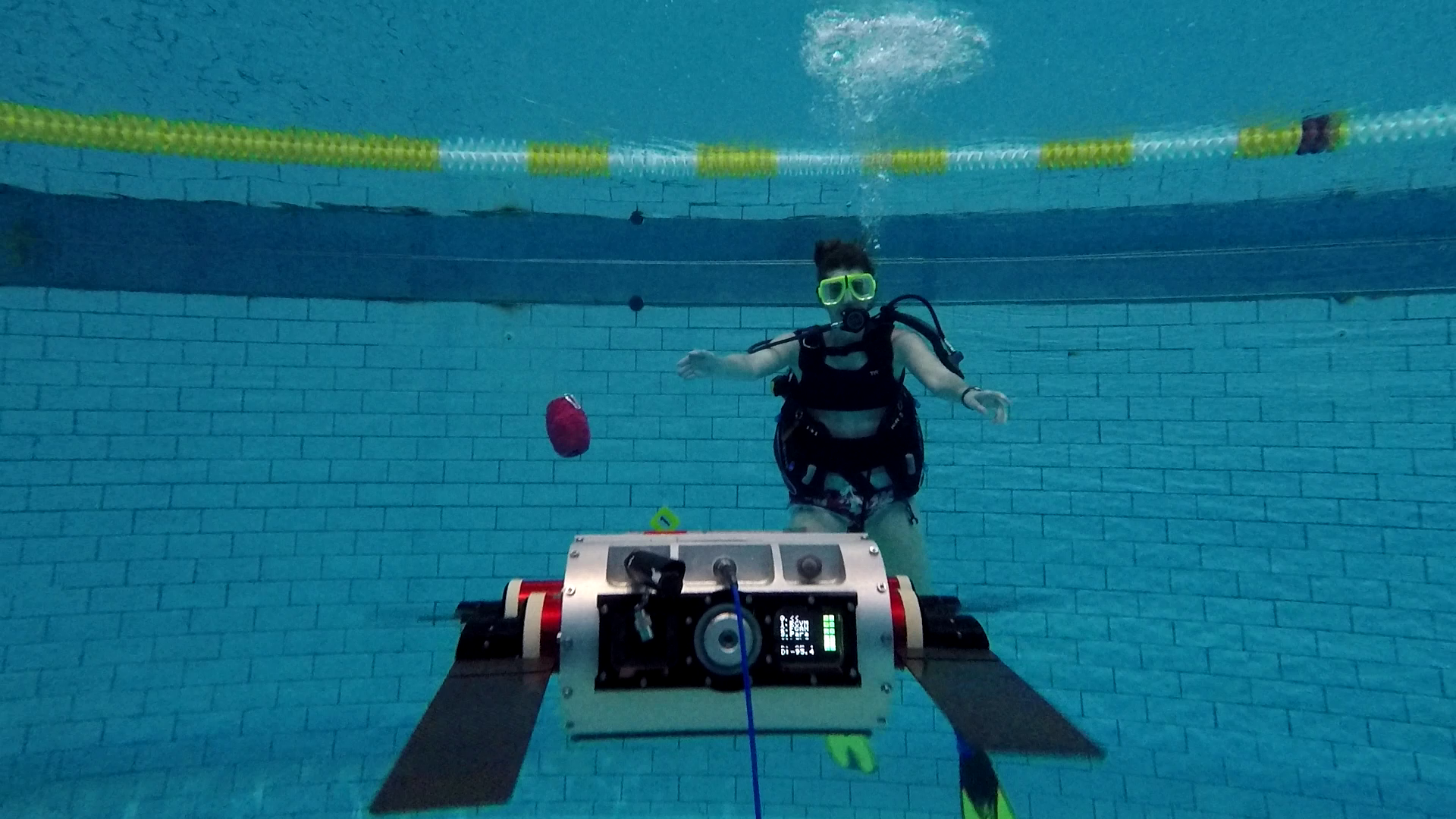}%
        \vspace{-1mm}
        \caption{Experimental scene of a closed-water test in which a diver points forward to an object. Through the use of DIP-3D, the Aqua AUV is able to locate the object.}%
    \label{fig:intro_img}
    \vspace{-4mm} 
\end{figure}

Many tasks in critical fields such as underwater infrastructure maintenance and biological monitoring often require divers to perform complex and even dangerous activities. Remotely operated vehicles (ROVs) have long been used to assist with these tasks, however the disconnect between operator and ROV is not ideal.
Due to limitations of field of view and tether connections, it is possible for the ROV to become entangled in the environment or overlook important information. 

Autonomous Underwater Vehicles (AUVs) are becoming more popular to perform tasks such as oil pipeline inspection~\cite{petillot_pipelone}, coral identification~\cite{ModasshirRobio2018_coralident} and archaeological surveys~\cite{bingham_archaeology}. 
Often, AUVs that are deployed to support these tasks are given hard-coded task information prior to deployment. 
Once deployed, \textit{in situ} mission reconfiguration is nearly impossible, since direct communication with AUVs is limited to acoustic communication channels \cite{stokey2005compact}. Acoustic communication is low-bandwidth and limits the complexity of potential information exchange.
Our goal is to assist in the development of a third paradigm (beyond `pure' ROV and AUV use) of underwater robotics in which experienced divers are able to work in collaboration with an \change AUV to \stopchange more safely and efficiently perform their tasks. AUVs such as  BUDDY~\cite{buddyAuv} and LoCO~\cite{loco_paper_2020} are designed with underwater human-robot interaction (U-HRI) in mind and can help bridge the gap between ROVs and \change fully autonomous AUVs.  \stopchange

Since diving is a cognitively demanding task \cite{fridman2018cognitive} and divers often manage many disparate objectives during a dive mission, bidirectional interaction between the AUV and human diver must be as natural and non-invasive as possible to prevent additional cognitive and physical burden to the diver. A variety of communication methods for U-HRI between divers and AUVs have been conceived. In a recent survey, Birk \etal~\cite{Birk2022_uhriSurvey} provides information about many of the challenges in U-HRI as well as state-of-the-art communication methods. Direct communication from diver to AUV is often vision-related, such as through the use of fiducial markers~\cite{Sattar2007CRV} or hand gestures performed by divers (\cite{Chavez_hri, caddy_gesture,islam_dynamic, Sattar2018JFR-Islam-MotionGestures}). Special tools such as dive gloves (\cite{caddy_gesture_dataset, audio_gloves}) can be  \change used to improve \stopchange communication as well.


Similar to scuba divers communicating an object or location through the use of shining lights or pointing physical objects such as sticks or poles, pointing gesture communication with an AUV relies on visual perception.
In \cite{Edge2023_DIP} it was shown that an RGB camera can be used to locate an area of interest to a diver and communicate this to an AUV within the two-dimensional image space; this is the first exploration of pointing gestures for communication with AUVs. 

The inclusion of distance in human-robot communication through pointing improves the ability to detect specific objects located in the direction of pointing when multiple objects are in the scene.
Underwater distance estimation overall is a challenging problem as significant distortions of incident visual light and optical attenuation negatively impact traditional distance estimation methods (see Fig. \ref{fig:stereo_underwater} for an example dense disparity map from an underwater diver scene) and sensors such as light-wave based Time-of-Flight RGB-D cameras become significantly impacted. In our work, we mitigate some challenges of dense stereo reconstruction by obtaining distance information for only points relevant to the pointing task through the use of sparse stereo triangulation.
\stopchange 

In this paper, we significantly improve 2D Diver Interest via Pointing~\cite{Edge2023_DIP} by combining pose-based gesture recognition with underwater stereo distance information to create the Diver Interest via Pointing in Three Dimensions (DIP-3D) algorithm. DIP-3D gives AUVs the ability to determine the position of an object pointed to by a diver in three dimensions \change(See Fig.~\ref{fig:intro_img})\stopchange. The contribution of this work is a modular visual framework for AUVs to determine the position of an object of interest pointed at by a human diver when there are multiple objects in the scene. We leverage 2D human body pose and object detection as well as sparse stereo reconstruction to provide scene distance information of an AUV to achieve this goal. 
 We also provide evaluations in the form of a human study and a closed-water \change experiments\stopchange.
\nopagebreak

\section{Background and related work}

\subsection{Gesture-based Communication with AUVs}
Interface for Diver-AUV communication through hand gestures is considered one of the primary communication vectors on deployed systems (\cite{Islam2018_DynamicReconfigure, Sattar2018JFR-Islam-MotionGestures, Chavez2021_UnderwaterVision-BasedGestureRecognition, caddy_gesture}). 
Of these,~\cite{caddy_gesture} includes directional hand signals and provides a stereo dataset for investigations into 3D algorithms.
In~\cite{Andrea_thesis} the diver's full body is used to signify that a diver is pointing through the use of deep-learning based object detectors (\eg Single Shot Multibox Detector (SSD)~\cite{liu2016_ssd}, Faster R-CNN~\cite{ren2017_fasterrcnn}). To provide location cues to an AUV, \cite{Edge2023_DIP} uses wrist and elbow keypoints of a pointing diver to indicate a 2D location of interest to an AUV.
We propose that including distance information when providing an AUV with locational cues can provide a more clear presentation of an area or object of interest to an AUV.

\subsection{Human Pose Applications}

\begin{figure}
    \centering
    \vspace{2mm}
    \includegraphics[width=.9\columnwidth]{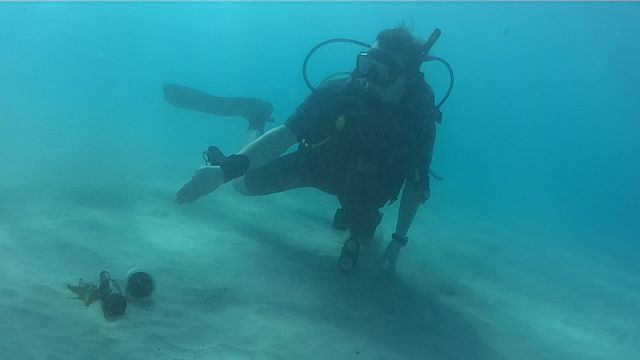}
    \caption{Example of a non-standard human pose typical during scuba diving operations. Diver-AUV interaction scenarios must accommodate these poses to be useful for underwater missions.}
    \label{fig:nonstandard_pose}
    \vspace{-2mm}
\end{figure}

Human pose estimators often locate landmarks, typically joints, on a human body which can then be used to provide information for pointing gestures (\cite{Tolgyessy_location, richarz_companion_loc,abidi_loc,Medeiros2021_fire}). 
Human pose estimation in both 2D and 3D is widely studied (see~\cite{CHEN2020Survey,Andriluka_2014_survey,SARAFIANOS2016_survey}). However, there does not exist a vision-based human pose landmark estimator designed for scuba divers. 
The use of human pose in underwater situations is gaining traction in the U-HRI community.
Chavez \etal~\cite{chavez_pose} uses point clouds of diver pose to perform diver tracking; however, specific pose landmarks are not found. 
Terrestrial-developed, monocular 2D pose estimators have been applied with some success either out-of-the-box or with some re-training; 
\eg Fulton \etal~\cite{fulton_adroc} use TRT pose in an autonomous diver approach algorithm~\cite{trtpose} and Islam \etal~\cite{Islam2021_pose} use OpenPose~\cite{openpose} to find relative positions of two robots when swimmers are in the water.

 Pose estimation networks often \change fail underwater \stopchange out-of-the-box for three reasons: 
\begin{enumerate*}
    \item Most frequencies of incident visual light are absorbed by seawater \cite{jackson1999classical}, leading to very different lighting conditions.
    \item Diver attire includes a host of life-support systems (\ie buoyancy control devices, breathing regulators, fins, wetsuits or drysuits).
    \item Poses a diver assumes underwater are less restricted than typical ambulation in the terrestrial domain (See Fig.~\ref{fig:nonstandard_pose}).
\end{enumerate*}
\stopchange

Our algorithm is estimator agnostic, and so any state-of-the-art \change method \stopchange could be used. We choose the Mediapipe~\cite{mediapipe} Pose~\cite{Bazarevsky_blazepose} framework, as in addition to low latency on CPU and impressive results on physical activities like yoga and dance, which also deal heavily with uncommon human poses, experiments in~\cite{Edge2023_DIP} show that this framework works for our purposes without additional training.

\subsection{Pointing to Provide Location}
 From rescue robotics~\cite{Delmerico_survey2.4} to pick-and-place actions~\cite{Shukla_ispointing,Littmann_Drees_Ritter_1996}, relaying location through the use of pointing gestures has seen success in multiple domains. 
Pointing pose gestures to communicate a location of interest have been used in terrestrial robotics for parking directives or exploration (\cite{Tolgyessy_location, richarz_companion_loc,abidi_loc, Azari_face_hand_realsense, van_explore}) and to indicate an object of interest on a shelf~\cite{Grossmann_Object}. In aerial robotics, a pointing pose can be used to direct a drone to a specific window~\cite{Medeiros2021_fire}.
In the underwater domain, pointing gestures have been used to locate an object of interest within an AUV's field of view~\cite{Edge2023_DIP}.
 The use of \change distance \stopchange when determining a pointing gesture target is a prevalent feature in communication through pointing. Only in~\cite{Edge2023_DIP, richarz_companion_loc} are the distances of the human or object not estimated. DIP-3D introduces the inclusion of target distance to select an object of interest in the underwater domain. 

\section{Methodology}
\label{sec:method}

\begin{figure}
\vspace{2mm}

    \centering
        \includegraphics[width=\linewidth]{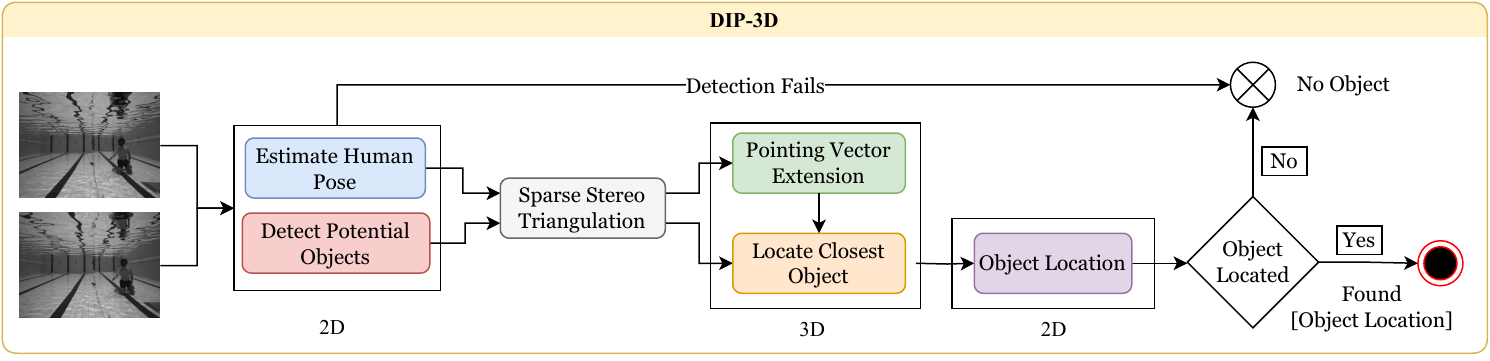}%
        \vspace{-1mm}
        \caption{Schematic diagram of the DIP-3D algorithm. Utilizing both the left and right images from the robot's stereo camera, we estimate 2D locations of both the human pose keypoints and candidate objects of interest. We then reproject these keypoints to 3D, where we discern the object of interest. Finally, we recover the object of interest in the 2D image plane for our visual servo control scheme. Colors in the diagram are coordinated with colors in Section~\ref{sec:evaluation}.}%
    \label{fig:DIP-3D_diagram}
    \vspace{-4mm} 
\end{figure}

To find objects of interests pointed to by a diver in 3D, we first determine the diver pose and potential objects of interest in two-dimensional image space $(x,y)$. Next, we perform sparse stereo triangulation to extract the corresponding three-dimensional points, $(x,y,z)$ so that we may extend a pointing vector and locate the nearest object. Finally, we recover the $(x,y)$ coordinates of the object of interest so that a controller in image space can be used by the AUV to move towards the object. We make the following assumptions:

\begin{itemize}
    \item The stereo camera has been calibrated. 
    \item The diver is situated in a pointing pose.
    \item The diver is pointing with their right arm, not pointing in front of or cross-body. The algorithm can be easily modified to use the left arm.
    \item \change The object of interest is present in both stereo pairs. \stopchange
\end{itemize}
\change 
A schematic of the DIP-3D algorithm can be seen in Fig.~\ref{fig:DIP-3D_diagram}.
\stopchange

\subsection{2D Diver Pose Estimation and Objects of Interest}
\label{sec:2d_estimation}
\change
Keypoint pose estimation of divers is an open problem both in 2D and 3D. We choose to begin in 2D, as there has been previous work with the use of 2D pose estimation and U-HRI. \stopchange
\change
As pose estimation is challenging, we minimize the number of body keypoints needed for our algorithm so that incorrect detections of unessential body parts will not inhibit use. We use the wrist, elbow, and shoulder as our keypoints. \stopchange
While only the wrist and elbow keypoints are required for pointing vector extension, we include the shoulder to filter infeasible 3D body poses caused by the challenge of underwater distance estimation.

Let $\bm{pose\_2D}$ define a set of 2D pose keypoints in image coordinates for an image of width $\text{W}$ and height $\text{H}$ which can be written for both the left and right stereo pairs as  
\begin{equation*}
    \bm{pose\_2D_{\text{left},\text{right}}} = \{({x}_{w}, {y}_{w}), ({x}_{e}, {y}_{e}), ({x}_{s}, {y}_{s})\},
\end{equation*}

\noindent \change where $x,y\in [0, \text{W}]\times[0, \text{H}]$, $w,~e,~\text{and}~s$ denote wrist, elbow, 
 and shoulder, respectively. \stopchange
Assuming a pose has been detected in both the left and right camera images, we proceed to identify potential objects of interest. Similar to~\cite{Edge2023_DIP}, \change we mask a portion of the images to prevent detection of objects within infeasible regions, based on our assumptions. \stopchange 
We mask the images from top to bottom, beginning at a constant distance from the right of the wrist to the left side of the image. We discard the region including the diver and anything to the left of the diver.

In principle, the object detection method depends on the actual mission given to the AUV and should be modified accordingly, \eg trash, coral, or artifact detection. \change For this work, since there is generally little background variation in the scene, and only salient features remain after masking, we use the low-level feature extractor \stopchange SIFT~\cite{Lowe04SIFT} for our detector.  In order to find matching keypoints between the images, we use brute force $k$-nearest neighbors matching. After finding keypoint matches, we use Lowe's ratio test as described in~\cite{Lowe04SIFT} with an empirically determined ratio of $0.3$ to discard features with poor matches. Let the candidate objects in the left and right images after filtering and masking be

\begin{equation*}
    \bm{obj\_2D}_{\text{left},\text{right}} = \{({x}_{0}, {y}_{0}), ..., ({x}_{n}, {y}_{n})\},
\end{equation*}

\noindent where $n$ denotes the number of candidate objects. We assemble the detected pose keypoints into a single set for each image as 
\begin{align*}
    \bm{k}_{\text{left}}&=\{\bm{pose\_2D}_{\text{left}},\bm{obj\_2D}_{\text{left}}\}\\
\bm{k}_{\text{right}}&=\{\bm{pose\_2D}_{\text{right}},\bm{obj\_2D}_{\text{right}}\}.
\end{align*}

\noindent \change
To discriminate between potential objects appearing at a different distance with respect to the AUV, we reproject these points to 3D space to identify the intended object of interest.
\stopchange

\subsection{Sparse Stereo Triangulation}
\label{sec:sparse_stereo}
\begin{figure}
     \centering
     \vspace{2mm}
     \begin{subfigure}[b]{0.3\columnwidth}
         \centering
         \includegraphics[width=\textwidth]{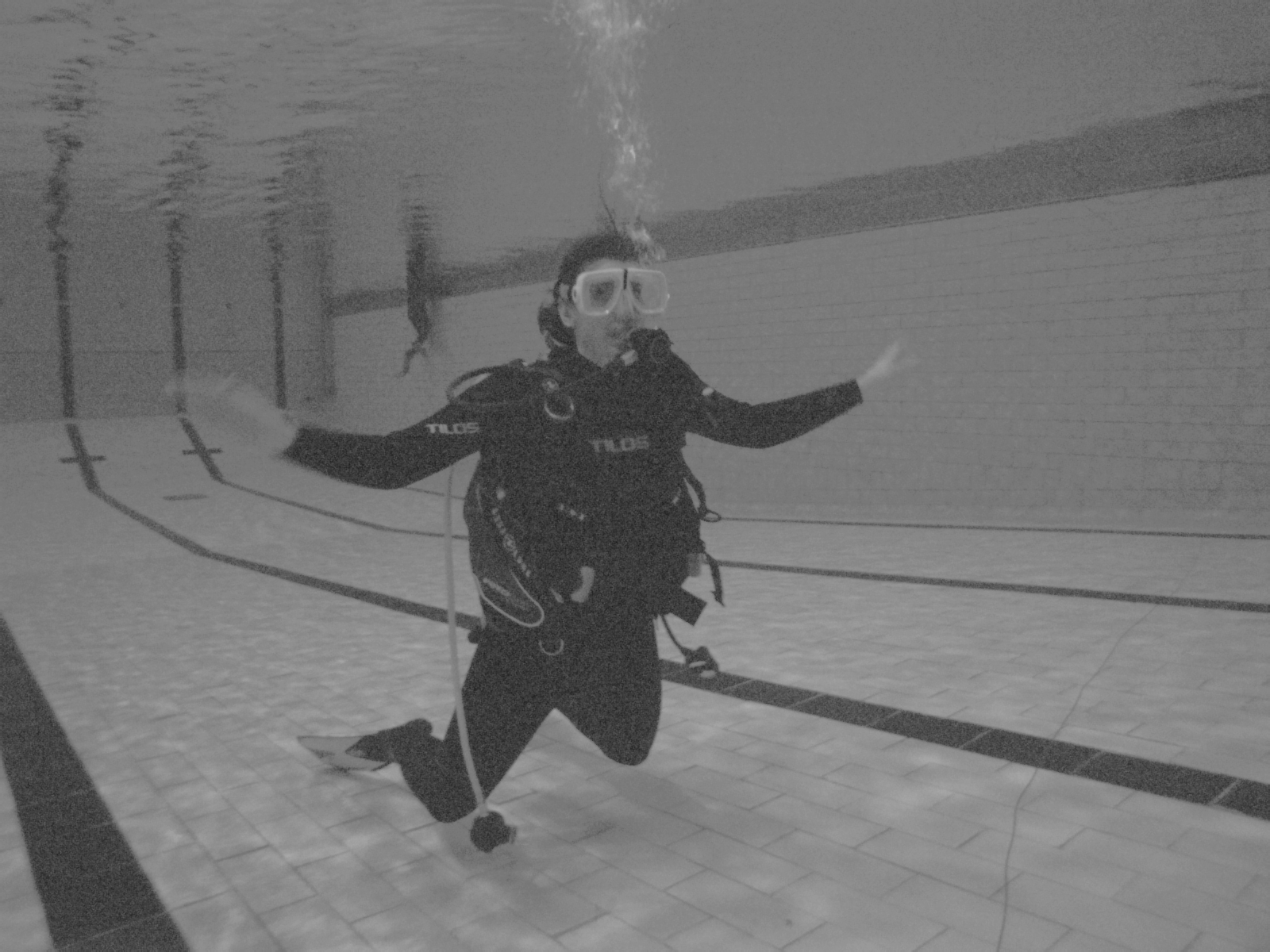}
         \caption{Left image.}
         \label{fig:cam_left}
     \end{subfigure}
     \hfill
     \begin{subfigure}[b]{0.3\columnwidth}
         \centering
         \includegraphics[width=\textwidth]{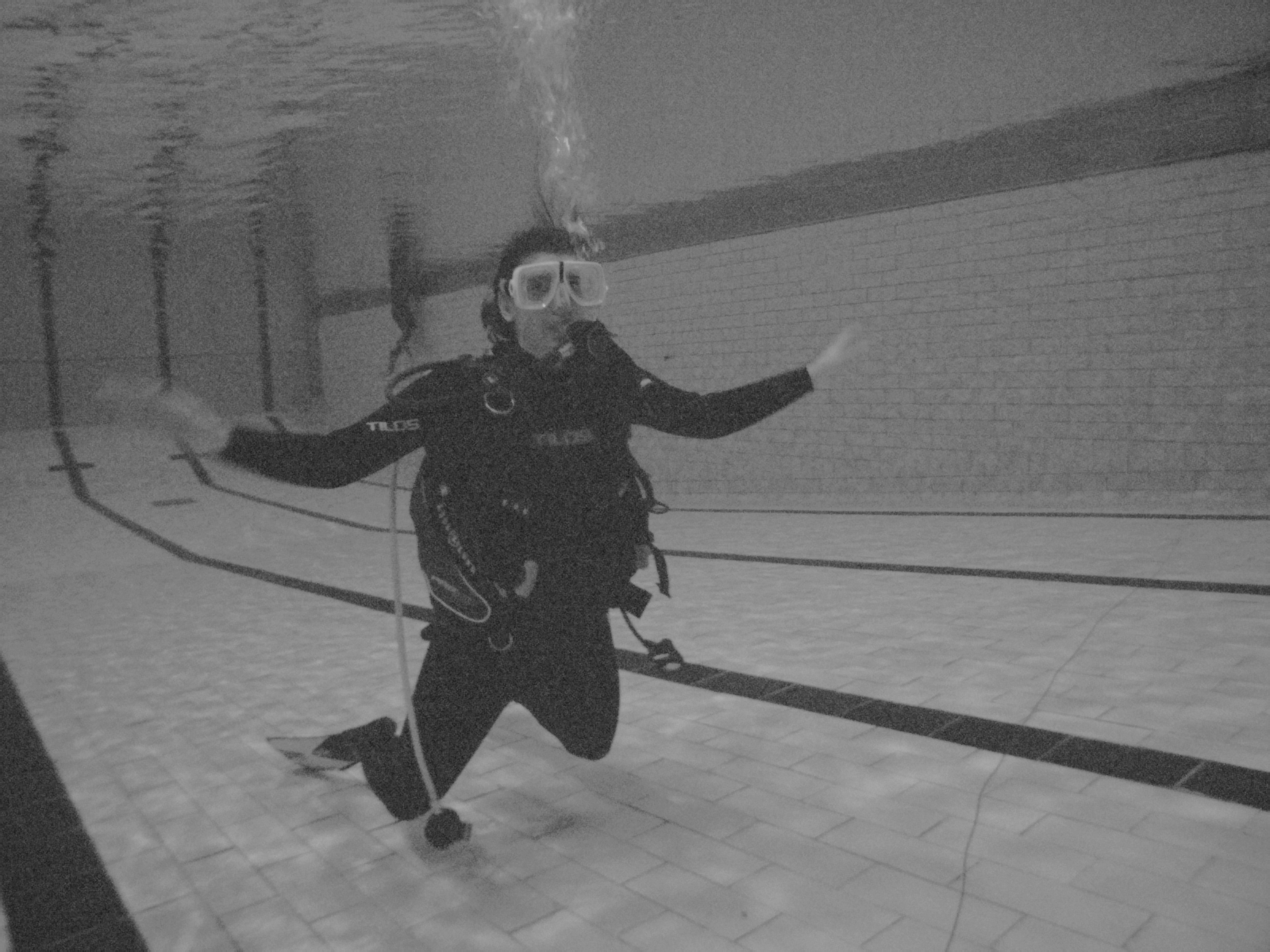}
         \caption{Right image.}
         \label{fig:cam_right}
     \end{subfigure}
     \hfill
     \begin{subfigure}[b]{0.3\columnwidth}
         \centering
         \includegraphics[width=\textwidth]{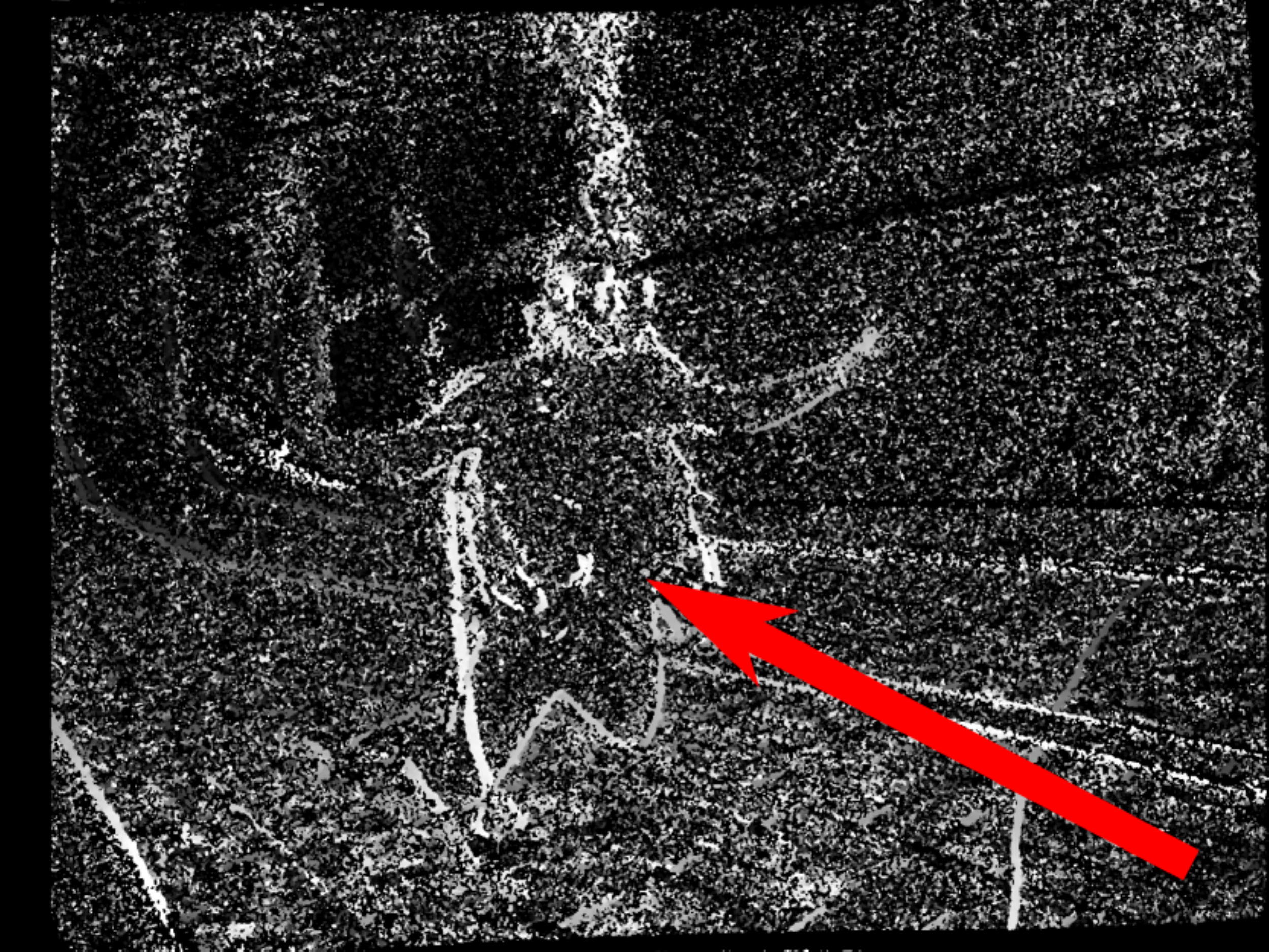}
         \caption{Disparity map.}
         \label{fig:example_disparity_map}
     \end{subfigure}
        \caption{Best if viewed at $175\%$ zoom. Demonstration of the challenges of traditional dense stereo reconstruction underwater using the Semi-Global Block Matching algorithm \cite{hirschmuller2007stereo}. Notice the disparity map contains inconsistencies within the diver's silhouette, shown by the red arrow.}
        \label{fig:stereo_underwater}
        \vspace{-1mm}
\end{figure}

Typical stereo correspondence algorithms such as Block Matching  and Semi-Global Block Matching (SGBM) \cite{hirschmuller2007stereo} algorithms rely on a cost function for computing stereo correspondence. This cost function is used to find the best match location between stereo image pairs for each pixel location. Effectively, the minimum cost matching location is chosen as the pixel match. While this works in the terrestrial domain, particularly in highly salient feature regions, computing this cost in the underwater domain leads to mismatches in correspondences. Fig.~\ref{fig:stereo_underwater} demonstrates an example disparity map output utilizing an OpenCV \cite{bradski2008learning} implementation of the SGBM algorithm. Notice the inconsistencies in illuminated pixels across the disparity map. This indicates that the algorithm finds many mismatches in corresponding pixels.
The accuracy of the pointing methodology in this work is highly coupled to precise distance estimation. To subvert the challenges we see with dense stereo reconstruction, we perform sparse triangulation by first computing the predicted pose in the camera left image. We then compute the predicted pose in the camera right image. We utilize triangulation to extract the camera frame three-dimensional points for the pose keypoints alone and not the entire scene. This process works as follows.

Let $(\bm{x}_{\text{left}},\bm{y}_{\text{left}})$ be the set of detected pose keypoints in the camera left image plane, and let $(\bm{x}_{\text{right}},\bm{y}_{\text{right}})$ be the detected set of keypoints in the camera right image. Assuming the images have been rectified using calibrated camera parameters so their epipolar lines are parallel, then we compute the disparity $D_i$  between the detected values in the horizontal coordinate $\bm{x}_{\text{left}}$ and $\bm{x}_{\text{right}}$ as 

\begin{equation}
    D_i=[\bm{x}_{\text{left}} - \bm{x}_{\text{right}}]_i = \frac{fB}{Z_{i}},
\end{equation}

\noindent where $f$ is the left image dominate camera focal length, $B$ is the baseline, and $Z_{i}$ is the distance to the pose keypoint $i$ in camera frame coordinates. We utilize the notation $Z_{i}$ to indicate an array element for every disparity value. We also need to recover the xy-coordinates of the pose keypoints to ensure proper extension of the pointing vector. We accomplish this by utilizing a reprojection matrix \cite{bradski2008learning} that operates on the vector set of image plane coordinates and disparity values. We do this for all pose keypoints with respect to the left image plane coordinates.

\begin{equation}
\begin{bmatrix}
    \bm{X}\\
    \bm{Y}\\
    \bm{Z}\\
    \bm{W}
\end{bmatrix}
=
\begin{bmatrix}
1&0&0&-c_{x}\\
0&1&0&-c_{y}\\
0&0&0&-f\\
0&0&-1/T_{x}&(c_{x}-c^{\text{right}}_{x})/T_{x}\\
\end{bmatrix}
\begin{bmatrix}
    \bm{x}_{\text{left}}\\
    \bm{y}_{\text{left}}\\
    \bm{D}\\
    \bm{1}
\end{bmatrix}.
\label{eqn:Q_comp}
\end{equation}

\noindent In (\ref{eqn:Q_comp}) we recover the camera frame three-dimensional locations scaled with respect to the homogeneous coordinate as $(x,y,z) = (\bm{X}/\bm{W}, \bm{Y}/\bm{W}, \bm{Z}/\bm{W})$. The values contained in the reprojection matrix are $T_x$, which is the translation component in the $x$-direction that translates from the left to the right image plane. The principal point components are $c_x$ and $c_y$ in the left image plane, and $c^{\text{right}}_x$ is the $x$-component of the principal point in the right image plane. 

After performing sparse stereo triangulation, we have the set of locations in 3D with respect to the left camera image. We define the three-dimensional pose locations as 
\begin{equation*}
    \bm{pose\_3D} = \{({x}_{w}, {y}_{w}, {z}_{w}), ({x}_{e}, {y}_{e}, {z}_{e}), ({x}_{s}, {y}_{s}, {z}_{s}\},
\end{equation*}

\noindent where $({x}, {y}, {z})\in\mathrm{R}^{3}$, 
and candidate object locations as 
\begin{equation*}
    \bm{obj\_3D} = \{({x}_{0}, {y}_{0}, {z}_{0}),..., ({x}_{n}, {y}_{n}, {z}_{n}\},
\end{equation*}

\noindent where $n$ is again the number of candidate objects found in 2D space. It is important to note that these 3D values are with respect to the camera frame and not a global or world frame.

\subsection{Choosing the Correct Object}
Before pointing vector extension and object selection, filters must be applied to the three-dimensional points to assist in limiting or removing incorrect information for the AUV to act upon. We first filter human pose information. 

We limit the difference in $z$ values between the wrist and elbow ($\text{mean} = 0.254$) and elbow and shoulder ($\text{mean}=0.377$) as we empirically determine the value should never be larger than $0.5$. \stopchange We also filter for invalid disparity computations which result in either $z<0$ or $z=\pm\infty$.

We are now left with reasonable 3D locations for the  diver's elbow and wrist along with potential objects of interest based on the AUV's left image. Let $\bm{w} = ({x}_{w}, {y}_{w}, {z}_{w})$ define the vector coordinates of the wrist keypoint, $\bm{e} = ({x}_{e}, {y}_{e}, {z}_{e})$ define the elbow keypoint, and $\bm{o}_i = ({x}_{i}, {y}_{i}, {z}_{i})$ define candidate object location, where $i\in[1,n]$. 
\begin{figure}[htbp]
    \centering
    \vspace{2mm}
    \includegraphics[width=.98\columnwidth]{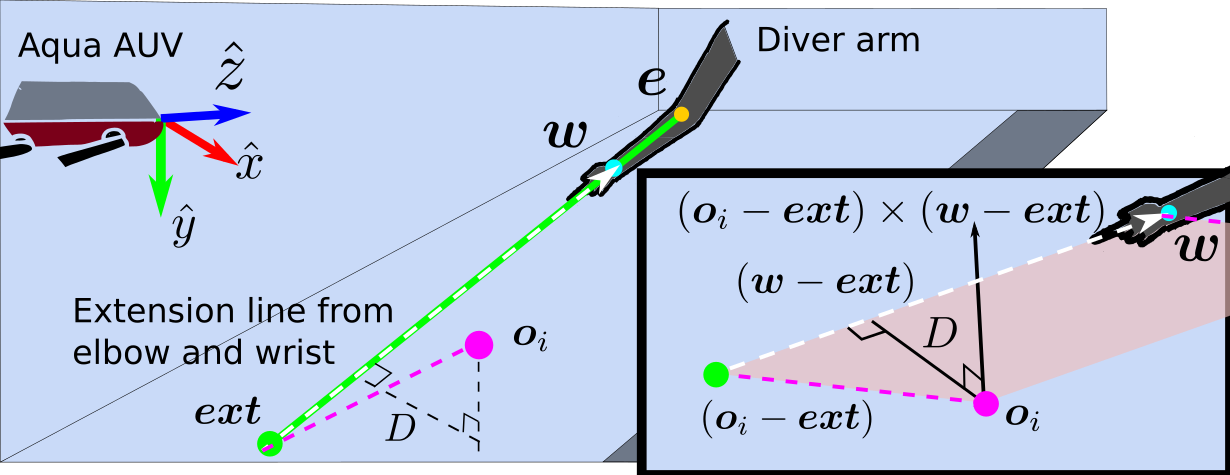}
    \caption{In the frame of reference of the robot's camera, we extend the line from the elbow to the wrist. Then we use 3D geometry to compute the perpendicular distance $D$ between each candidate object and the extended pointing line.}
    \label{fig:pointing_idealization}
    \vspace{-1mm}
\end{figure}
In order to determine the most probable object of interest, we first project a line $\bm{ext}$ in the direction extending from the elbow, through the wrist, and towards the direction of pointing as 
\begin{equation} \label{eq:point_extension}
\centering
  \begin{aligned}
    \bm{ext} = \bm{w} + \text{s}_{\text{f}} * (\bm{w}-\bm{e}),
  \end{aligned}
\end{equation}
where $\text{s}_{\text{f}}$ is a scaling factor that can be modified as needed.
We use an empirically determined scaling factor $\text{s}_{\text{f}}=3$, because $\text{s}_{\text{f}}$ is the approximate distance between the camera frame origin and the wrist point. This ensures the extension line intersects the $xy$-plane, which ensures we can recover objects that appear in the foreground of the robot's image plane.



\change 
To find the object of interest, we find the object $o^{*}_i\in\bm{obj\_3D}$ that minimizes the perpendicular distance 
 $D_i$ between the object and the line that connects the extension and wrist. This process is demonstrated in Fig.~\ref{fig:pointing_idealization}. From 3D geometry, we know

\begin{equation}
    D_i = \frac{\lVert (\bm{o}_i - \bm{ext})\times (\bm{w}-\bm{ext})\rVert_{2}}{\lVert\bm{w}-\bm{ext} \rVert_{2}}.
\end{equation}
\stopchange

\change We recover the 2D object location using the index of the minimum object distance, since our control method works on 2D image coordinates.  \stopchange
\nopagebreak
\section{Experimental Setup and Evaluations}
\label{sec:evaluation}

\begin{figure}
     \centering
    \vspace{2mm}
     \begin{subfigure}[b]{0.48\columnwidth}
         \centering
         \includegraphics[width=\textwidth]{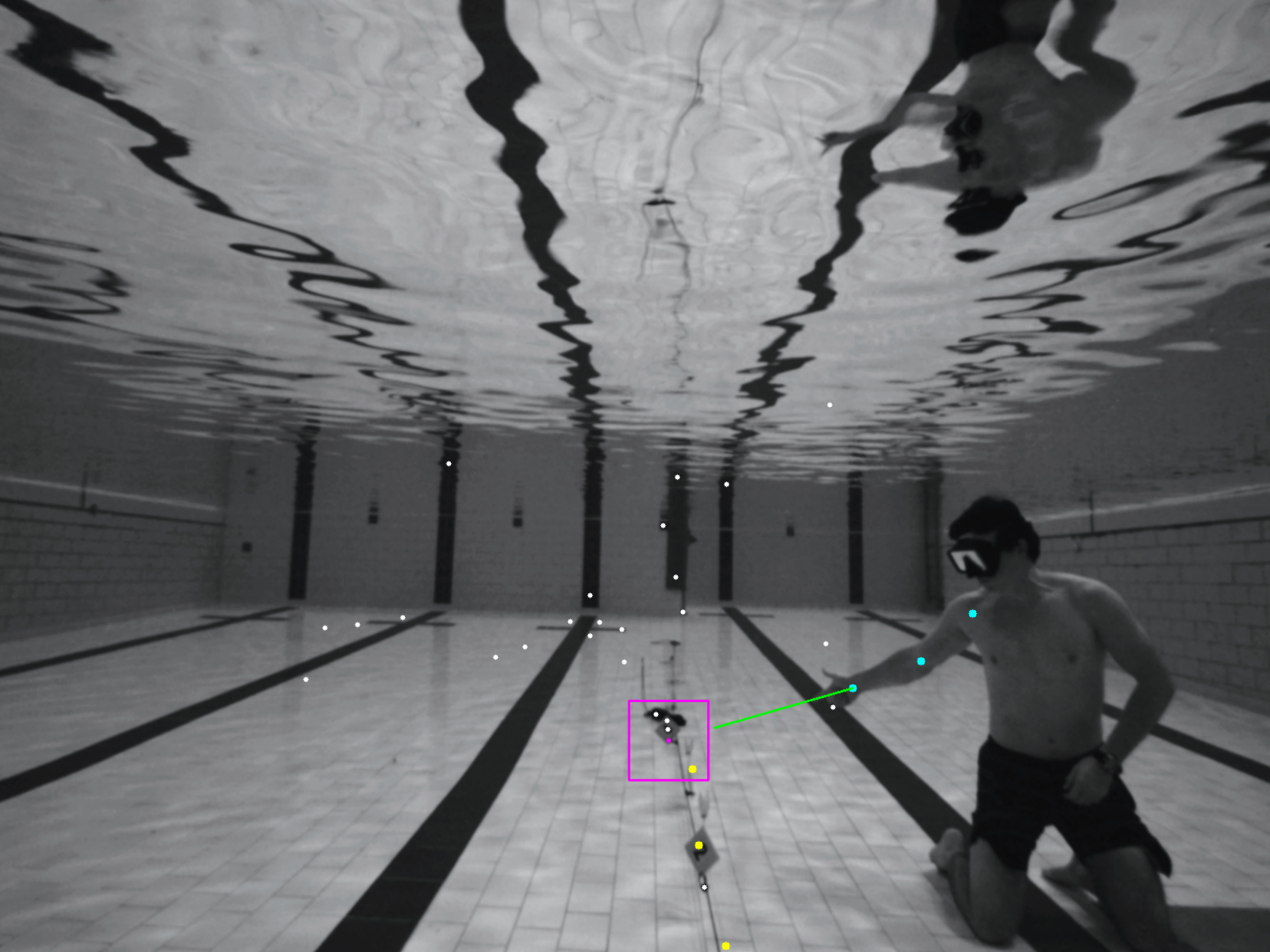}
         \caption{}
         \label{fig:point_back}
         
     \end{subfigure}
     \hfill
     \begin{subfigure}[b]{0.48\columnwidth}
         \centering
         \includegraphics[width=\textwidth]{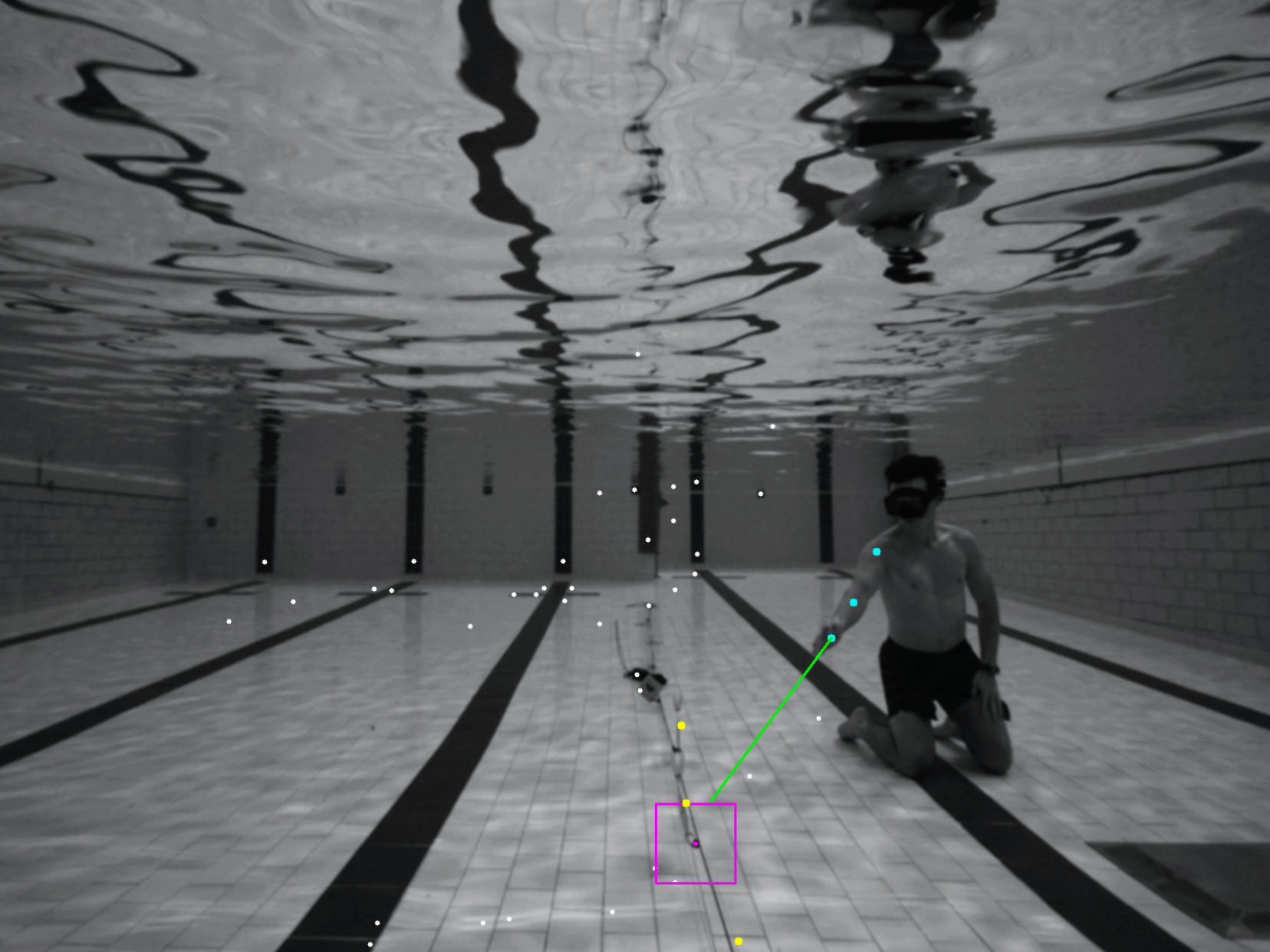}
         \caption{}
         \label{fig:point_front}
     \end{subfigure}
     \caption{\change Two examples of images from the DIP-3D distance evaluation, the diver is pointing away from and then towards the AUV. In each case, DIP-3D is able to accurately locate the direction of the point. The green vector denotes the pointing direction projected back to 2D space and white points denote other potential objects. Images enhanced to aid readability, best seen at 150\% zoom\stopchange.}
        \label{fig:distance_eval_example}
\end{figure}
The objective of DIP-3D is to locate an object or the direction the diver is pointing in three dimensions. There are \change several challenges in evaluating our algorithm: 
\begin{enumerate*}
\item Obtaining precise ground truth measurements for object locations is challenging due to the constant relative motion of the robot, diver, and object.
\item Multiple objects must be located within the AUV's image space while the diver is pointing, with the approximate locations of these objects in relation to each other known as well. 
\item The direction the diver is pointing during the data collection phase must be known, as it can be difficult to determine the pointing locations in the resulting images.
\end{enumerate*}\stopchange 

\begin{figure}
    \centering
    \includegraphics[width=.980\columnwidth]{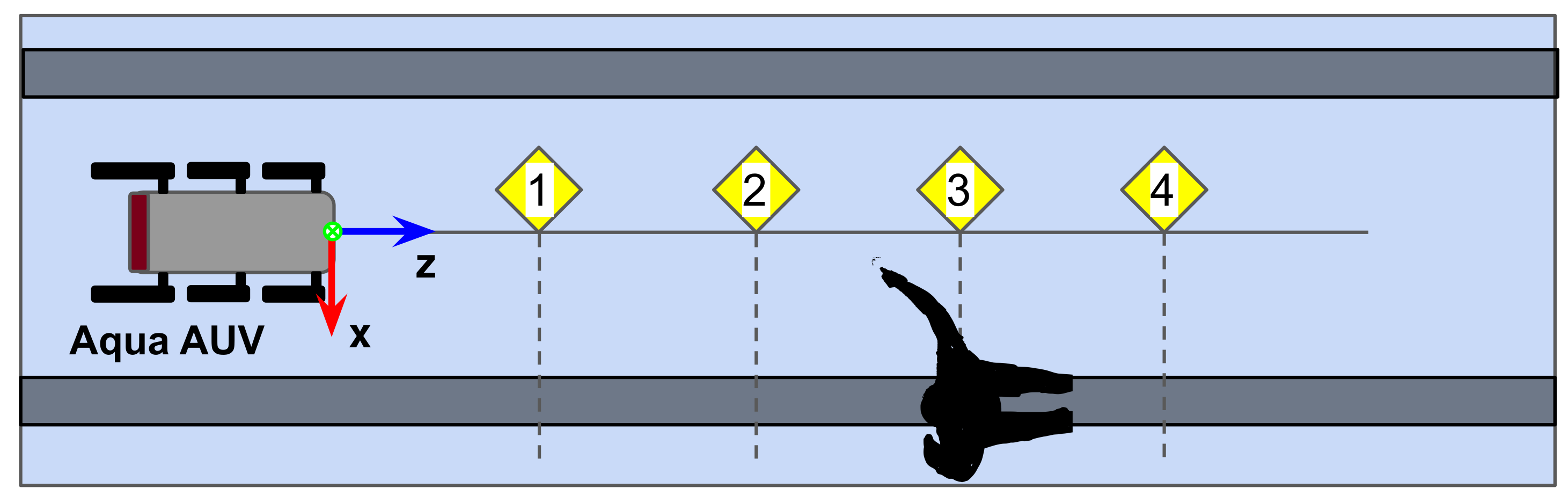}
    \caption{Top view of the closed-water experimental setup. A track line with meter markings was positioned along the center line between two lane markers. The human was positioned along the right lane line from the robot's perspective.}
    \label{fig:experimental_setup}
    \vspace{-3mm}
\end{figure}

In order to minimize these challenges, we utilized a partially instrumented setup, shown in Fig.~\ref{fig:experimental_setup}, where a track line was maintained between two lane markers in a closed-water environment. The track line delineated meter distances from the robot to objects of interest, showing an approximate distance from the diver to the AUV's camera.
As everything was free-floating, these numbers \change provided reasonable approximations to the ground truth\stopchange. We calibrated our stereo camera system using an AprilTag calibration target \cite{richardson2013aprilcal}.
Proper calibration has drastic effects on the quality of 3D reconstruction \cite{Shortis2019_calibration, lavest2000underwater_calibration}, so we collected ample calibration data and ran \textit{in situ} calibration of our stereo cameras.
With this information, we now show that the AUV is able to discern the direction (near or far) of the object of interest \change from the AUV. Specific situations are discussed during the different evaluations\stopchange.

\change
\subsection{DIP-3D Directional Evaluation}
\stopchange
\begin{table*}[t]
\vspace{2mm}
\renewcommand{\arraystretch}{1.2}
\caption{Euclidean distance error between ground truth labeled images and predicted object locations when the correct object location matches the predicted.}
\begin{center}
\newcolumntype{R}{>{\centering\arraybackslash}X}%
\begin{tabularx}{\textwidth}{l |R|R|R|R}
\toprule
&\textbf{Pointing at Object 1}&\textbf{Pointing at Object 2}&\textbf{Pointing at Object 3}&\textbf{Error (pixels)}\\
    \cline{1-5}
\multicolumn{1}{l|}{\textbf{Person Position 2}} &$65.1\pm68.5~\bf{(24)} $&$141.9\pm122.34~\bf{(42)} $&$188.59\pm144.4~\bf{(7)} $&$131.86\pm50.91$\\ 
\multicolumn{1}{l|}{\textbf{Person Position 3}} &-&$116.7\pm153.03~\bf{(16)} $&$87.47\pm32.64~\bf{(9)} $&$102.09\pm14.62$\\
\cline{1-5}
\multicolumn{1}{l|}{\textbf{Error (pixels)}} & $65.1\pm0.0$&$129.3\pm12.6$&$138.03\pm50.56$&$119.95\pm43.05$\\ 
    \bottomrule
\end{tabularx}
\end{center}
\footnotesize{$^b$ Errors are reported in \textit{mean}$\pm$\textit{standard deviation (observation count)} format in pixel units.}\\
\label{table:total_classification_accuracy}
\end{table*}

\begin{figure*}
\vspace{-7mm}
    \centering
    \begin{subcaptionblock}{0.19\textwidth}
        \centering
        \includegraphics[width=\textwidth]{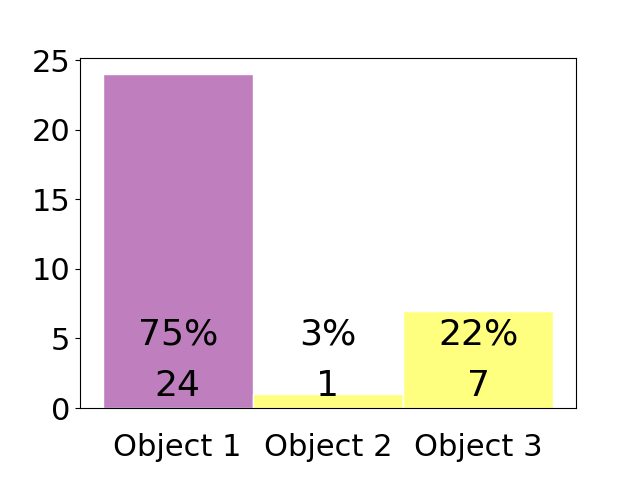}
        \captionsetup{justification=centering}
        \caption{Person position 2, pointing at position 1}
        \label{fig:2_1_hist}
    \end{subcaptionblock}
    \begin{subcaptionblock}{0.19\textwidth}
        \centering
        \includegraphics[width=\textwidth]{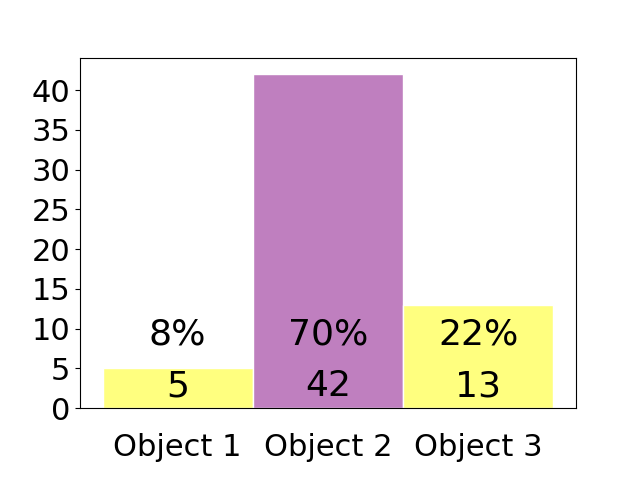}
        \captionsetup{justification=centering}
        \caption{Person position 2, pointing at position 2}
        \label{fig:2_2_hist}
    \end{subcaptionblock}
    \begin{subcaptionblock}{0.19\textwidth}
        \centering
        \includegraphics[width=\textwidth]{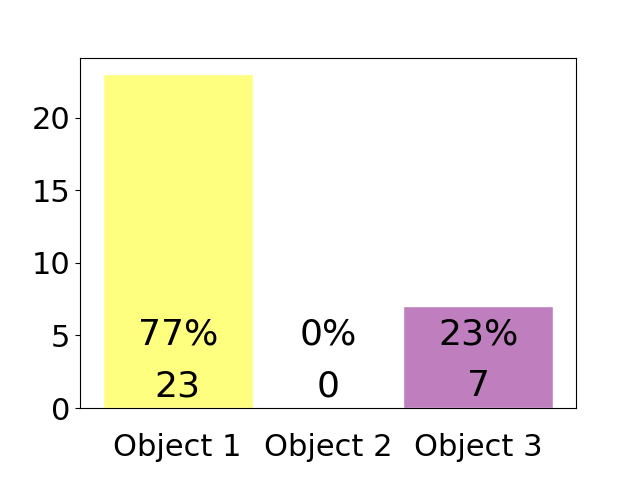}
        \captionsetup{justification=centering}
        \caption{Person position 2, pointing at position 3}
        \label{fig:2_3_hist}
    \end{subcaptionblock}
    \begin{subcaptionblock}{0.19\textwidth}
        \centering
        \includegraphics[width=\textwidth]{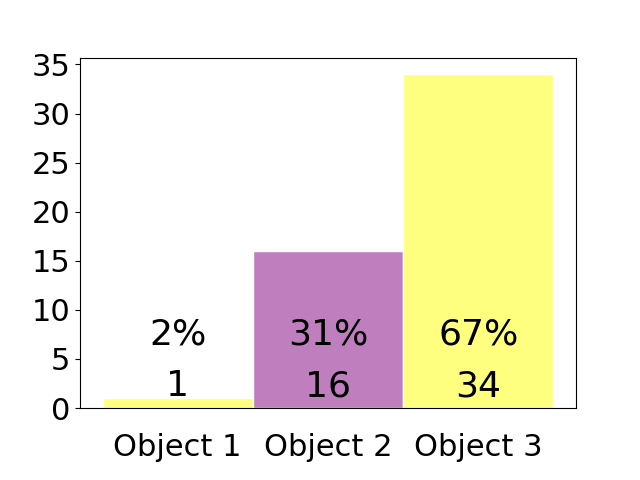}
        \captionsetup{justification=centering}
        \caption{Person position 3, pointing at position 2}
        \label{fig:3_2_hist}
    \end{subcaptionblock}    
    \begin{subcaptionblock}{0.19\textwidth}
        \centering
        \includegraphics[width=\textwidth]{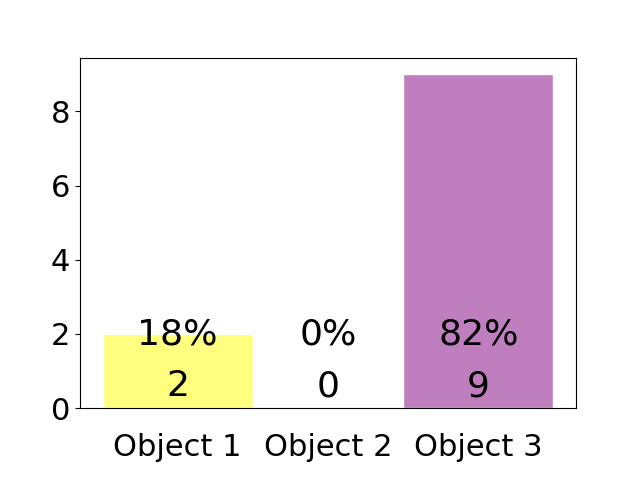}
        \captionsetup{justification=centering}
        \caption{Person position 3, pointing at position 3}
        \label{fig:3_3_hist}
    \end{subcaptionblock}       
    \caption{Comparison of the relative frequency that the detection system predicts the ground truth object at which the person is pointing during the in-water testing. Purple histogram bars indicate the ground truth object. Yellow indicates some other object.}\label{fig:histogram_results}
\vspace{-6mm}
\end{figure*}

Images for this evaluation were recorded in a ROS~\cite{ROS2009ICRA} bagfile and extracted as image pairs with an image size of $1600 \times 1200$. Images with the desired object and pointing diver in frame were retained. Of those, each image pair included in this work contains a detected pointing diver and at least one detected feature. It is not required that a possible feature exists on or near the desired target in the images. During recording of the bagfile, each pose and object location was held for \change approximately $6$ seconds. \change  The quality of pose estimation and feature detection in the stereo pair varies by the distance from the AUV, so the number of valid pairs of detected objects in each location varies\stopchange. It is important to note the challenges different distances present\stopchange.

The images were rectified and each of the number placards $1$,~$2$,~and~$3$ were annotated with a 2D point. \change The evaluated distance from the AUV to the diver and object is limited by the detection results on the rectified images (\ie the pose of the diver was undetectable farther than 3 meters)\stopchange.
Fig. \ref{fig:histogram_results} shows a breakdown of the count of times the location of the `closest object' was nearest to the intended target. The closer the object and diver to the AUV, the more accurate the results. The inaccuracies in the farther distance are due to a combination of a lack of potential objects near the target as well as a lack of defined poses for the diver. This is especially clear when both the person and object are located about $3$ meters from the AUV, where a total of only $11$ images in which a pose and at least one object was found.  
Table \ref{table:total_classification_accuracy} shows the total Euclidean distance error of the between the located object and the ground truth label when the predicted object is the closest. When pointing direction and the nearest object are in agreement, the average distance in pixels between the two is $119.95 \pm 43.05$. \change Samples of the images used for this evaluation can be found in Fig.~\ref{fig:distance_eval_example}. In addition to the yellow ground truth points, we include white points denoting all potential features. The sparsity of matching features can be seen in the foreground of Fig.~\ref{fig:point_back}. We also include a 2D  projection of the 3D pointing direction as the green line segment to show that the object chosen is in the direction of the pointing gesture. Additional samples are presented in the accompanying video\stopchange.

\change
\subsection{DIP-3D and Human-based Target Location Correlation}
\begin{figure}
     \centering
     \vspace{1mm}
     \begin{subfigure}[b]{0.48\columnwidth}
         \centering
         \includegraphics[width=\textwidth]{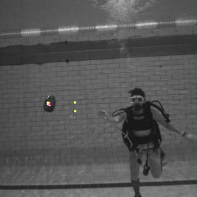}

     \end{subfigure}
     \hfill
     \begin{subfigure}[b]{0.48\columnwidth}
         \centering
         \includegraphics[width=\textwidth]{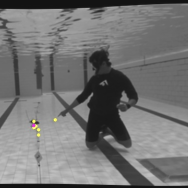}

     \end{subfigure}
     \caption{Selected images from the human survey. Projected annotations are shown in yellow, DIP-3D results in purple. Images enhanced and cropped for readability.}
        \label{fig:human_study}
        \vspace{-5mm}
\end{figure}

\stopchange
A human study was performed to provide a reference on the difficulty of locating an object in 3D space given a 2D image. Ten images were chosen in which DIP-3D accurately identified an object in the direction the diver was pointing. Each image includes a pointing diver and the \change potential objects\stopchange. Nine participants were asked to draw a four-sided bounding box on the RGB color image where they believed the diver was pointing. Post-processing allows us to project the centroid of the bounding box to the same frame of reference as the 2D detected object found by our algorithm.
The average pixel distance from annotated centroid to our result is $45.9\pm60.28$. An example of the rectified image results is shown in Fig.~\ref{fig:human_study}, in which purple marks the predicted object and yellow marks the human annotations. As can be seen, the added dimension of distance perception assists greatly in locating the result of a pointing gesture. A closer look at this survey, including a breakdown of the images and their annotations are presented in the accompanying video.

\subsection{Runtime}
With the current experimental setup, using Mediapipe Pose as the backbone along with the SIFT algorithm, DIP-3D \change runs at $0.857$ \stopchange fps on an Intel\textsuperscript{TM} i7-6700 CPU, which is acceptable for AUV operations. For each pair of images, both Mediapipe and SIFT run twice, once on each image.

\section{Challenges}

As seen in the bench experiments, we are limited in distance by the accuracy of stereo reconstruction. It is possible that using a wider camera baseline would improve distance results, however that \change presents different logistical challenges, particularly in the dimensions of the companion AUV\stopchange. A further challenge is the detection of incorrect pose of a diver when they are pointing in-front-of or cross-body as an incorrect detection will lead to an incorrect pointing location. However, to improve this, the pose estimator itself would need to be greatly improved. 

\section{Conclusion}

We introduce scene distance information to direct an AUV to a three-dimensional location of interest. Although the use of 3D algorithms in the underwater environment presents many unique  challenges, we show that
an AUV is able to locate a chosen object in the direction the diver is pointing when multiple objects are in the field of view. When the diver is within mission-specific distance of the AUV, it is highly likely that the direction estimated through sparse stereo triangulation agrees with the direction the diver is pointing. Future work will improve communication through the inclusion of gesture classification for downstream robotic tasks, as well as the design of a feedback method from the AUV to reduce ambiguity of objects of interest.
\bibliographystyle{IEEEtran}
\bibliography{references}

\end{document}